\documentclass[a4paper, 10pt, conference]{ieeeconf}      

\IEEEoverridecommandlockouts                              

\usepackage{comment}
\usepackage{tabularx}
\usepackage{booktabs}
\usepackage{subcaption}
\usepackage{caption} 
\usepackage{mathtools}
\usepackage{amsmath}    
\usepackage{amssymb}    
\usepackage{graphicx}   
\usepackage{bm}         
\usepackage{url}
\usepackage{graphicx}      

\usepackage{xcolor}
\usepackage{float}      
\usepackage{multirow}
\usepackage{bbm}
\usepackage{booktabs} 
\usepackage{extdash}
\usepackage{multirow}
\usepackage{array}     
\usepackage{boldline}  
\usepackage{siunitx}
\let\labelindent\relax
\usepackage{enumitem}
\usepackage{tikz} 
\usetikzlibrary{shapes, arrows.meta}

\usepackage[sorting=none]{biblatex}
\usepackage{caption}
\title{\LARGE \bf
COM-PACT: COMponent-Aware Pruning for Accelerated Control Tasks in Latent Space Models
}

\author{Ganesh Sundaram$^{*}$$^{1}$, Jonas Ulmen$^{*}$$^{1}$, Amjad Haider$^{*}$$^{1}$ and Daniel Görges$^{*}$$^{1}$
\thanks{*Authors have contributed equally}
\thanks{$^{1}$Department of Electrical and Computer Engineering, RPTU University Kaiserslautern-Landau, Germany. E-mail:
        {\tt\small \{ganesh.sundaram, jonas.ulmen, ahaider, daniel.goerges\}@rptu.de}}
}

\begin{document}

\maketitle
\thispagestyle{empty}
\pagestyle{empty}

\begin{abstract}

The rapid growth of resource-constrained mobile platforms, including mobile robots, wearable systems, and Internet-of-Things devices, has increased the demand for computationally efficient neural network controllers (NNCs) that can operate within strict hardware limitations. While deep neural networks (DNNs) demonstrate superior performance in control applications, their substantial computational complexity and memory requirements present significant barriers to practical deployment on edge devices. This paper introduces a comprehensive model compression methodology that leverages component-aware structured pruning to determine the optimal pruning magnitude for each pruning group, ensuring a balance between compression and stability for NNC deployment. Our approach is rigorously evaluated on Temporal Difference Model Predictive Control (TD-MPC), a state-of-the-art model-based reinforcement learning algorithm, with a systematic integration of mathematical stability guarantee properties, specifically Lyapunov criteria. The key contribution of this work lies in providing a principled framework for determining the theoretical limits of model compression while preserving controller stability. Experimental validation demonstrates that our methodology successfully reduces model complexity while maintaining requisite control performance and stability characteristics. Furthermore, our approach establishes a quantitative boundary for safe compression ratios, enabling practitioners to systematically determine the maximum permissible model reduction before violating critical stability properties, thereby facilitating the confident deployment of compressed NNCs in resource-limited environments.

\end{abstract}

\section{INTRODUCTION}
\label{sec:introduction}

The integration of neural networks into control systems represents a paradigm shift from classical methods, offering unparalleled adaptability and performance, especially for complex nonlinear systems. NNCs excel at learning from high-dimensional observational data, such as raw pixels, and can effectively compress this information into powerful latent space representations. However, a fundamental conflict arises when attempting to integrate these advanced controllers into real-world applications. The significant computational and memory resources demanded by NNCs are often incompatible with the stringent constraints of the embedded hardware found in automotive systems, robotics, and IoT devices. This gap between capability and deployability presents a critical barrier to system integration.

Model compression techniques, particularly pruning, are essential for bridging this implementation gap. However, while these methods have been successfully applied in domains like computer vision, their direct application to NNCs presents unique and substantial challenges. A fundamental disconnect exists between the objectives of standard compression and the requirements of control theory. Compressing a predictive model, such as an image classifier, primarily risks a drop in accuracy. In contrast, compressing a controller risks the stability and safety of a physical system, where failures can have catastrophic real-world consequences. This high-stakes domain demands a new approach to compression that looks beyond mere accuracy.

The foremost requirement for any controller is stability, which is often established through rigorous mathematical formulations, such as those based on the Lyapunov stability theory. Standard pruning techniques, which alter network weights and structure, can inherently invalidate these formal guarantees. In a closed-loop system, even small, compression-induced output deviations can accumulate over time, leading to oscillatory, unpredictable, or hazardous system behavior. Therefore, for the successful system integration of NNCs, any compression method must treat control-related properties not as secondary concerns but as primary constraints.

This paper presents a framework for resource-focused pruning aimed at enabling the safe and efficient deployment of NNCs on embedded hardware. Building upon our prior work in application-specific pruning~\cite{sundaram2025applicationspecific}, we extend the methodology to incorporate control-theoretic constraints. In particular, we formulate a novel pruning problem in which maintaining Lyapunov stability is treated as a primary constraint, alongside strict targets for model size and inference time. To validate the proposed approach, we demonstrate its effectiveness on a complex NNC architecture and investigate the use of post-pruning fine-tuning to recover performance lost during compression, further improving the feasibility of real-world system integration.

\section{Related Works}
\label{sec:related_works}

Neural network pruning has evolved significantly from early heuristic approaches to sophisticated, theoretically grounded frameworks. Early pruning strategies relied on simple magnitude-based approaches, which remove weights with the smallest absolute values. Despite their simplicity, these methods have proven surprisingly effective, often matching or surpassing more complex techniques~\cite{guptaComplexityRequiredNeural2022}. However, the limitations of unstructured pruning, particularly the lack of practical speedup on standard hardware, led to the development of structured pruning methods. 

Structured pruning addresses these limitations by removing entire parameter groups, directly reducing computational complexity while maintaining dense operations, essential for edge device deployment~\cite{heStructuredPruningDeep2023}. The field has evolved beyond magnitude-based criteria to incorporate sophisticated methods, including Hessian-based importance estimation, class-aware strategies~\cite{chongResourceEfficientNeural2023, jiangClassAwarePruningEfficient2023}, and operator-theoretic frameworks unifying magnitude and gradient approaches~\cite{redmanOperatorTheoreticView2022}. To counter performance degradation, researchers integrate complementary techniques such as knowledge distillation~\cite{qianBoostingPrunedNetworks2024}, sparse optimization with regularization~\cite{shiSparseOptimizationGuided2024b}, gradient-based methods, and pruning-quantization combinations~\cite{wangDifferentiableJointPruning2020}. Energy-constrained optimization techniques further enable pruning within specified energy budgets~\cite{chengSurveyDeepNeural2024b, liuEnergyConstrainedOptimizationBasedStructured2022}.

Structured pruning aligns well with hardware acceleration through compiler optimizations~\cite{gongAutomaticMappingBestSuited2021} and platforms like HALP that balance accuracy with latency constraints~\cite{shenStructuralPruningLatencySaliency2022}. Reported performance improvements are substantial: $2\times$ acceleration for YOLOv7 on FPGA~\cite{pavlitskaIterativeFilterPruning2024}, $19.1\times$ speedup for LSTM on Jetson platforms~\cite{lindmarIntrinsicSparseLSTM2022}, and $2\times$ improvement for GNNs~\cite{gurevinPruneGNNAlgorithmArchitecturePruning2024b}. Additional optimizations include compute-in-memory architectures with $11.1\times$ size reduction~\cite{mengExploringComputeinMemoryArchitecture2022b} and systolic array enhancements achieving $4.79\times$ speedup~\cite{zhangSCRASystolicFriendlyDNN2023b}. However, traditional methods rely on architecture-specific grouping strategies, limiting applicability. While SPA offers architecture-agnostic pruning~\cite{wangStructurallyPruneAnything2024}, DepGraph~\cite{fangDepGraphAnyStructural2023f} provides a more comprehensive solution through dependency graph analysis and norm-based criteria, enabling flexible and consistent pruning across diverse architectures.

While these frameworks address architectural generalization, a critical gap remains in how pruning affects the functional and structural interdependencies of different network components, particularly in multi-component architectures where different modules serve distinct purposes. This limitation is especially pronounced in control applications, where maintaining system stability and safety properties is critical, a challenge that existing pruning methods do not adequately address.

\section{Neural Network Controllers on Embedded Hardware}
\label{sec:NNConEmbeddedHardware}

The successful deployment of an NNC on resource-constrained hardware presents a dual challenge. The controller must not only operate within the stringent physical limitations of the hardware but also rigorously adhere to the principles of control theory to ensure safe and reliable operation. These two distinct sets of requirements, one driven by the hardware's resource budget and the other by the system's control-theoretic properties, must be jointly satisfied for successful system integration. The following subsections elaborate on both categories of constraints.

\subsection{Resource-Related Constraints}
\label{subsec:resource_constraints}

The successful integration of NNCs into physical systems is fundamentally constrained by the hardware limitations inherent to embedded platforms. These devices are designed to prioritize cost, power efficiency, and a small physical footprint over raw computational power. Consequently, any NNC intended for deployment must adhere to a strict set of resource budgets. The primary constraints include:

\vspace{0.1em} 
\noindent\textbf{Memory:} Embedded systems operate with stringent memory limitations, both for volatile memory (RAM), which is often in the range of kilobytes to a few megabytes, and for non-volatile storage, which dictates the maximum size of the controller's program code.

\vspace{0.1em} 
\noindent\textbf{Computation:} The processing power, often measured in Floating Point Operations Per Second (FLOPs), is significantly lower than in desktop systems. This directly impacts inference time (latency), which is a critical performance metric.

\vspace{0.1em} 
\noindent\textbf{Real-Time Performance:} Many control applications, particularly in safety-critical domains, have hard real-time requirements, meaning they must execute tasks within a guaranteed timeframe. This necessitates careful optimization and often relies on a Real-Time Operating System (RTOS).

\vspace{0.1em} 
\noindent\textbf{Power and Energy:} Power consumption is a first-order design constraint. Firmware and hardware are heavily optimized to minimize energy use, often employing low-power sleep modes to extend battery life.

These resource limitations are hard requirements that dictate the feasibility of deploying an NNC. Therefore, they must be treated as primary design constraints in any practical system integration workflow.

\subsection{Control-Theoretic Constraints}
\label{subsec:control_constraints}

When pruning an NNC, performance cannot be measured solely by task-specific metrics, such as reward or accuracy. Unlike general predictive models, NNCs are subject to rigorous control-theoretic requirements that govern the safety and reliability of a physical system. The fundamental among these is stability, a property ensuring that a system's output remains bounded and predictable. This is often formally verified using mathematical frameworks, such as the Lyapunov stability theory or contraction theory. Furthermore, properties such as sensitivity and robustness are critical for ensuring a controller can handle model uncertainties and external disturbances. Therefore, any viable pruning strategy for NNCs must treat the preservation of these control-related characteristics.

\subsection{TD-MPC Inference}
\label{sec:tdmpcinference}

To quantitatively assess the computational and deployment challenges posed by NNCs, we selected the TD-MPC agent, designed to balance an inverted pendulum directly from raw pixel observations. The model architecture and hyperparameters follow those identified as optimal benchmarks for control performance~\cite{hansen2022temporaldifferencelearningmodel}. The characteristics of the trained TD-MPC model are detailed in Table~\ref{tab:model_params}, including input representation, model capacity, and achieved control performance.

\begin{figure*}[ht]
    \captionsetup{width=\textwidth}
    \centering
    \includegraphics[width=\textwidth]{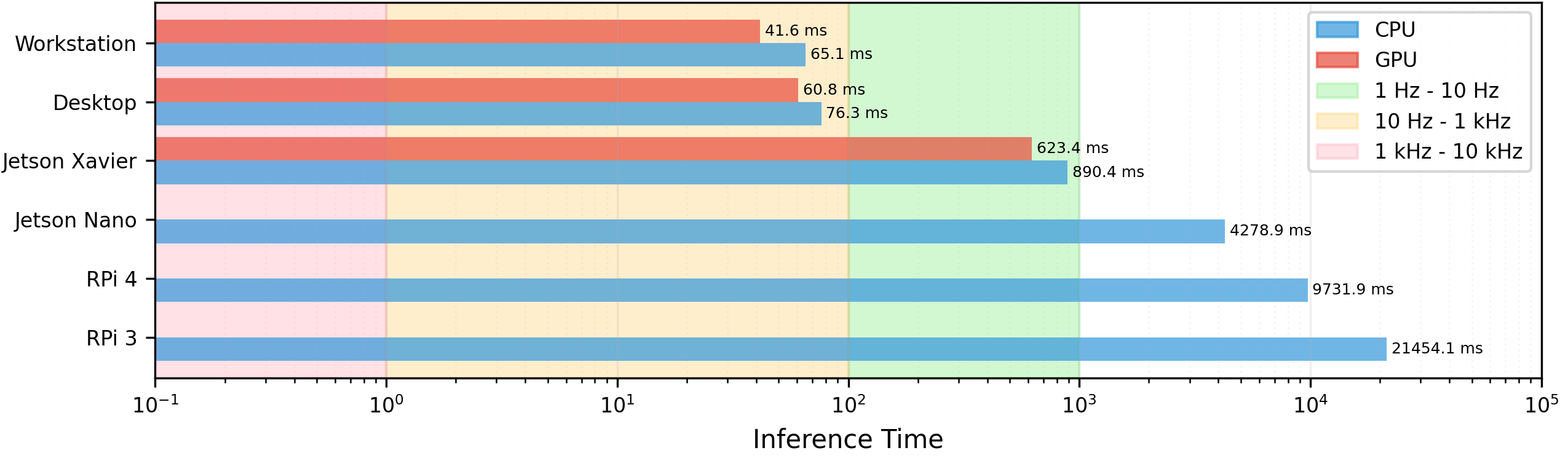}
    \caption{Inference time benchmark of the uncompressed TD-MPC model on various hardware. The colored regions represent typical real-time operating frequencies for control systems: slow dynamics (green), mid-range applications like robotics (yellow), and high-speed systems (red).}
    \label{fig:all_hw_inference_tdmpc}
\end{figure*}

To evaluate real-world deployability, the agent's inference performance was benchmarked across a spectrum of hardware platforms, from high-performance workstations to resource-constrained embedded systems, including the Jetson Nano and Raspberry Pi series. For control systems like the inverted pendulum, where controller frequencies must range between \SI{10}{Hz} and \SI{1}{kHz} (yellow region in Fig.~\ref{fig:all_hw_inference_tdmpc}), the results reveal a substantial performance gap between platform categories. The inference times on all tested embedded systems are orders of magnitude higher than on high-end desktop hardware. This significant computational bottleneck demonstrates that the direct deployment of the uncompressed model is infeasible, firmly establishing model compression as a prerequisite for successful system integration.

\vspace{-0.5em}
\begin{table}[ht]
\centering
\caption{}
\label{tab:model_params}
    \begin{tabular}{@{}lc@{}}
        \toprule
        \textbf{Parameter}          & \textbf{Value} \\ \midrule
        Input Image Size            & $28 \times 28$ pixels \\
        Network Components          & Encoder, Dynamics, Reward, Policy \\
        Layers per Component        & 3 \\
        Hidden Units per Layer      & 512 \\
        Latent Dimension            & 50 \\
        Episode Length (steps)      & 125 \\
        Episode Reward (avg.)       & 858.6 \\
        Total Parameters            & 1,548,360 \\
        Total FLOPs per Inference   & $4.12 \times 10^{8}$ \\
        \bottomrule
        \bottomrule
    \end{tabular}
\end{table}

\vspace{-1em}
\section{Resource-Aware Structured Pruning}
\label{sec:pruning_methodology}

Given the sensitivity of multi-component NNCs and the need to preserve control-theoretic properties, a targeted compression strategy is essential. Unlike global techniques such as quantization that can unpredictably alter the entire model, pruning carefully eliminates redundant elements, making it the preferred method for this application. Specifically, structured pruning is chosen over unstructured pruning, which creates inefficient sparse matrices. This technique eliminates entire functional blocks \cite{frankleLotteryTicketHypothesis2019}, preserving the architectural integrity necessary for stable control while producing a dense, compact model that operates efficiently on embedded hardware.

\subsection{Component-Aware Pruning Groups}
\label{subsec:pruning_groups}

A prerequisite for effective structured pruning is the partitioning of a model's parameters into groups of structurally and functionally coupled elements. Given the multi-component nature of the TD-MPC architecture, we employ the Component-Aware DepGraph method from~\cite{sundaram2025enhancedpruningstrategymulticomponentarXiv} to perform this segmentation.

This approach analyzes the model's computational graph to identify both component-specific groups (parameters exclusive to one component) and coupling groups (parameters shared between components). The analysis yields a total of ten distinct pruning groups: eight component-specific and two coupling groups, which form the basis for our pruning strategy. Table~\ref{tab:pruning_groups} provides a detailed breakdown of these groups, enumerating their constituent modules and parameter counts.

\vspace{-1.5em}
\begin{table}[ht]
\caption{}
\label{tab:pruning_groups}
\centering
\scriptsize
\sisetup{group-separator={,}}
    \begin{tabular}{@{}llccc@{}}
        \toprule
        \textbf{Group Type} & \textbf{Component} & \textbf{Group} & \textbf{Modules} & \textbf{Parameters} \\ \midrule
        \multirow{12}{*}{\textit{Comp.-specific}} & 
            \multirow{3}{*}{Encoder} & 1 & 1 & \num{25632} \\ & & 2 & 1 & \num{9248} \\ & & 3 & 1 & \num{9248} \\ \cmidrule(l){2-5} & 
            \multirow{1}{*}{Dynamics} & 1 & 1 & \num{262656} \\ \cmidrule(l){2-5} & 
            \multirow{1}{*}{Reward} & 1 & 1 & \num{262656} \\ \cmidrule(l){2-5} &
            \multirow{1}{*}{Pi} & 1 & 1 & \num{262656} \\ \cmidrule(l){2-5} &
            \multirow{1}{*}{Q1} & 1 & 1 & \num{262656} \\ \cmidrule(l){2-5} &
            \multirow{1}{*}{Q2} & 1 & 1 & \num{262656} \\   \midrule
        \multirow{3}{*}{\textit{Coupling}} & 
             \multirow{1}{*}{Encoder-Pi} & 1 & 2 & \num{26112} \\ \cmidrule(l){2-5} &
             \multirow{1}{*}{EncoderPi-Dyn.Rew.Q1Q2} & 1 & 5 & \num{369152} \\ 
        \bottomrule
        \bottomrule
    \end{tabular}
\end{table}

\vspace{-1.5em}
\subsection{Pruning Coefficient Search}
\label{subsec:coefficient_search}

To precisely control the level of sparsity applied to each of the ten identified groups, we introduce a set of corresponding pruning coefficients. Each group $i$ is assigned a coefficient \(c_i \in [0, 1]\), which dictates the fraction of parameters to be removed from that specific group. The complete pruning configuration for the controller is therefore defined by the coefficient vector \(\mathbf{c} = [c_1, c_2, \dots, c_{10}]^T\).

The major challenge is to find an optimal vector \(\mathbf{c}\) that compresses the TD-MPC model sufficiently to meet the resource constraints of the target embedded system (Section~\ref{subsec:resource_constraints}) while simultaneously preserving critical control-theoretic properties, such as stability (Section~\ref{subsec:control_constraints}). To solve this complex, multi-objective challenge, we formulate the search for these coefficients as a formal optimization problem.

\subsection{Performance Evaluation Function}
\label{subsec:perfevalfunction}

To rigorously assess the stability properties of both the original and pruned TD-MPC agents, a dedicated neural Lyapunov function~\cite{10015199, chang2019neural} was developed for the inverted pendulum environment with the encoded latent spaces as its input. Training is carried out such that the Lyapunov function $V$ is constrained to be positive definite for all non-equilibrium states, with $V=0$ enforced at the upright equilibrium position. Moreover, the learning objective ensures that $\Delta V$, the change in $V$ as the system evolves, is negative definite whenever the agent is moving toward equilibrium, reflecting the fundamental requirements for Lyapunov stability.

\begin{figure}[ht] 
    \captionsetup{width=\linewidth}
    \centering
    \includegraphics[width=\linewidth]{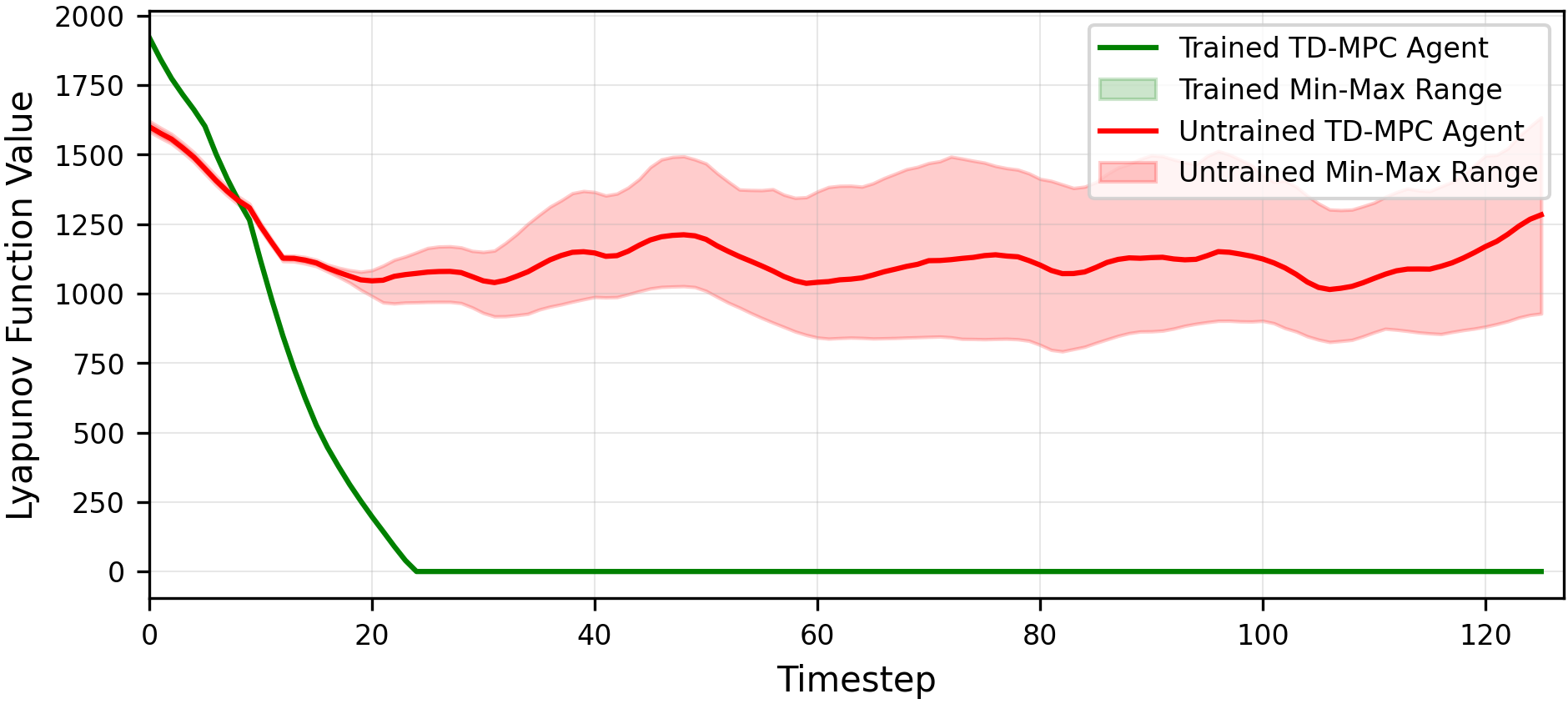}
    \caption{Mean Lyapunov function values across five evaluated episodes for untrained (bold red) and trained (bold green) TD-MPC agents. Shaded areas indicate the min-max Lyapunov function value ranges observed for each agent.}
    \label{fig:lyapTDMPC}
\end{figure}

After successful training, the neural Lyapunov function is used to evaluate the closed-loop behavior of the parent TD-MPC agent. During multiple episodes in which the agent balances the pendulum, the output of the Lyapunov function is monitored. From Fig.~\ref{fig:lyapTDMPC}, it is evident that the value $V$ initially starts at a high positive value and decreases monotonically, reaching zero. The convergence to zero closely matches the observed settling time of the pendulum in simulation, verifying the accuracy of the trained Lyapunov network. Furthermore, once the system has stabilized, $V$ remains at zero for the remainder of the episode, demonstrating that the TD-MPC agent can maintain the pendulum in its upright position and that the Lyapunov function reliably captures this stable state. With the neural Lyapunov function established as our stability verification framework, we can now define the optimization objective for our resource-aware pruning strategy. Unlike traditional application-specific pruning approaches that optimize for performance metrics such as reward~\cite{sundaram2025applicationspecific}, our optimization-based coefficient search prioritizes preserving the monotonic decrease in the Lyapunov function as the system evolves toward equilibrium. This formulation ensures that stability becomes the primary constraint driving the pruning process. 

\vspace{-0.5em}

\vspace{-1.5em}
\section{Experiments and Discussion}
\label{sec:experiments}

To evaluate whether the proposed resource-aware pruning strategy can facilitate the integration of NNCs into embedded platforms, we designed a two-stage experimental framework. The first stage targets the practical challenge of reducing model size to fit available hardware resources, while the second stage explores the fundamental limits of compression.

\vspace{0.5em}
\noindent\textbf{Experiment I:} \textit{Achieving a Sparsity Target:} The initial goal is to validate that our method can achieve a precise, modest level of compression without violating stability. The optimization is constrained to a target model sparsity \(\rho=10\%\) with a tolerance of \(\varepsilon=+1\%\), meaning the final sparsity must fall within the range of $[10\%, 11\%]$.

\vspace{0.2em}
\noindent\textbf{Experiment II:} \textit{Determining the Stability Boundary:} The second stage aims to identify the maximum possible compression the TD-MPC agent can withstand before its stability is compromised. In this scenario, the sparsity constraint is removed from the optimization problem. The sole objective is to maximize the pruning ratio while ensuring the Lyapunov stability condition (i.e., the monotonic decrease of \(V\)) is never violated.

\begin{figure*}[ht]
    \captionsetup{width=\textwidth}
    \centering
    \includegraphics[width=0.95\textwidth]{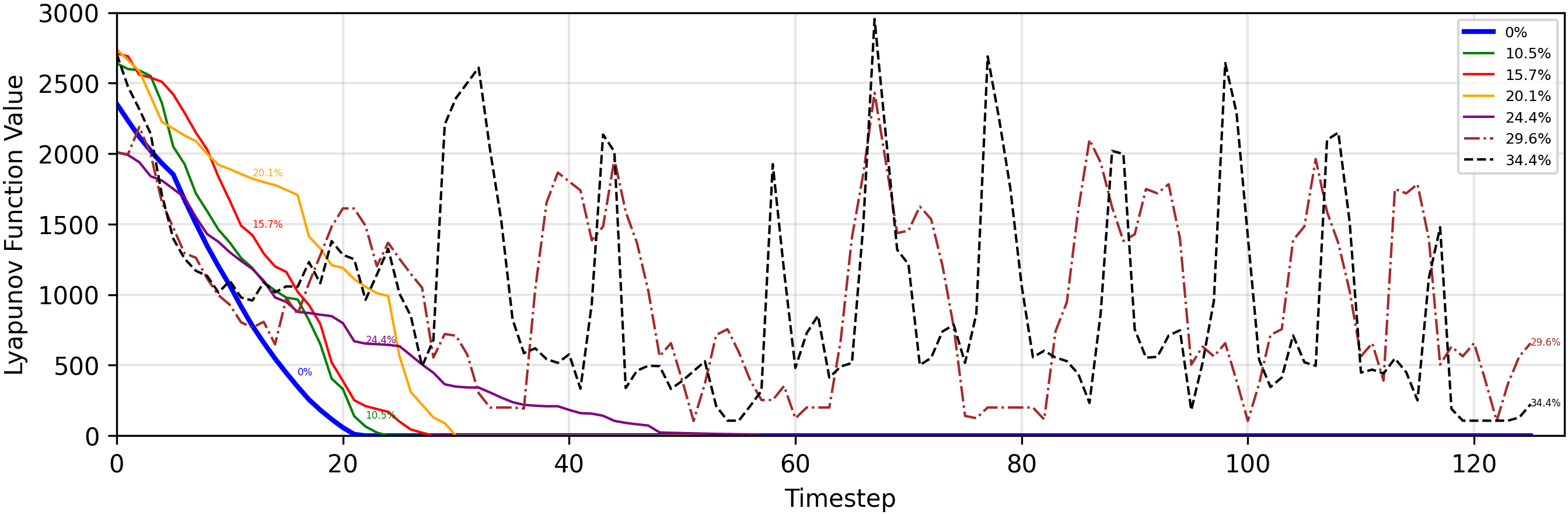}
    \caption{Lyapunov function evolution across different sparsity levels of the TD-MPC agent. The unpruned baseline (blue) demonstrates ideal stability with monotonic convergence to zero. Progressive pruning from 10.5\% to 20.1\% maintains stability with gradually increasing settling times. A critical transition occurs around 24.4\% sparsity, beyond which the system exhibits unstable oscillatory behavior (29.6\% and 34.4\%), clearly violating the Lyapunov stability condition.}
    \label{fig:lyapvalues}
\end{figure*}

The results from both experiments are visualized in Fig.~\ref{fig:lyapvalues}, which reveals critical insights into the relationship between pruning aggressiveness and system stability. The analysis demonstrates three distinct behavioral regimes that validate our approach. The unpruned baseline model (blue line) exhibits ideal Lyapunov behavior, with V values starting around 1900 and monotonically decreasing to zero by timestep 22. Remarkably, our method maintains this fundamental stability characteristic across all compression levels up to 20.1\% sparsity. The pruned models at 10.5\%, 15.7\%, and 20.1\% all preserve the essential monotonic decrease property, successfully converging to zero despite modest increases in settling time, demonstrating that substantial model compression, reducing the original model by one-fifth, can be achieved while maintaining rigorous stability guarantees.

At 24.4\% sparsity, the system exhibits the first clear signs of stability degradation, with non-monotonic reduction in V values and thus energy and prolonged settling time, representing a critical threshold where the controller could potentially begin to lose its stability margin. Beyond 25\% sparsity, the system experiences catastrophic stability failure. Both the 29.6\% and 34.4\% pruned models exhibit highly erratic, oscillatory Lyapunov function trajectories that fail to converge to zero, showing persistent energy oscillations that violate the fundamental Lyapunov stability condition. The severity of oscillations increases with pruning aggressiveness, confirming fundamental controller instability. These results establish a safe compression sweet spot of approximately 22\% sparsity, where substantial model reduction is achieved without compromising stability. The transition from stable to unstable behavior occurs rapidly around the 24-25\% threshold, emphasizing the importance of staying within safe compression limits. Table~\ref{tab:pruningcoeff} details the final pruning coefficient configurations for both component-specific and coupling groups that achieved the specified sparsity levels, along with their corresponding episode rewards. Notably, coupling groups were deliberately excluded from pruning (coefficients = 0) due to their inherent sensitivity arising from their complex inter-component dependencies. This optional strategic decision not only preserves critical architectural connections but also reduces the complexity of the optimization search space, allowing the algorithm to focus on component-specific parameter allocation.    

\vspace{-0.8em}
\begin{table}[ht]
\centering
\caption{}
\label{tab:pruningcoeff}
\resizebox{\linewidth}{!}{%
    \begin{tabular}{@{}c c c c@{}}
        \toprule
        \textbf{Sparsity}   & \textbf{Component-Specific Group Coefficients}            & \textbf{Coupling} & \textbf{Reward}\\ \midrule
                    0       & [0.0, 0.0, 0.0, 0.0, 0.0, 0.0, 0.0, 0.0]                  & [0, 0]            & 858.6 \\
                    10.5    & [0.0, 0.0, 0.0, 0.0, 0.190, 0.196, 0.102, 0.112]          & [0, 0]            & 851.4 \\
                    15.7    & [0.105, 0.095, 0.0, 0.275, 0.198, 0.142, 0.131, 0.159]    & [0, 0]            & 804.3  \\
                    20.1    & [0.0, 0.115, 0.114, 0.386, 0.453, 0.106, 0.124, 0.159]    & [0, 0]            & 793.7  \\
                    24.4    & [0.124, 0.0, 0.113, 0.584, 0.465, 0.104, 0.159, 0.114]    & [0, 0]            & 689.3  \\
                    29.6    & [0.097, 0.051, 0.020, 0.641, 0.623, 0.149, 0.099, 0.116]  & [0, 0]            & 633.1  \\
                    34.4    & [0.114, 0.116, 0.0, 0.726, 0.708, 0.276, 0.124, 0.159]    & [0, 0]            & 602.1  \\
        \bottomrule
        \bottomrule
    \end{tabular}%
}
\end{table}

\vspace{-1.5em}
\begin{figure}[htbp] 
    \captionsetup{width=\linewidth}
    \centering
    \includegraphics[width=\linewidth]{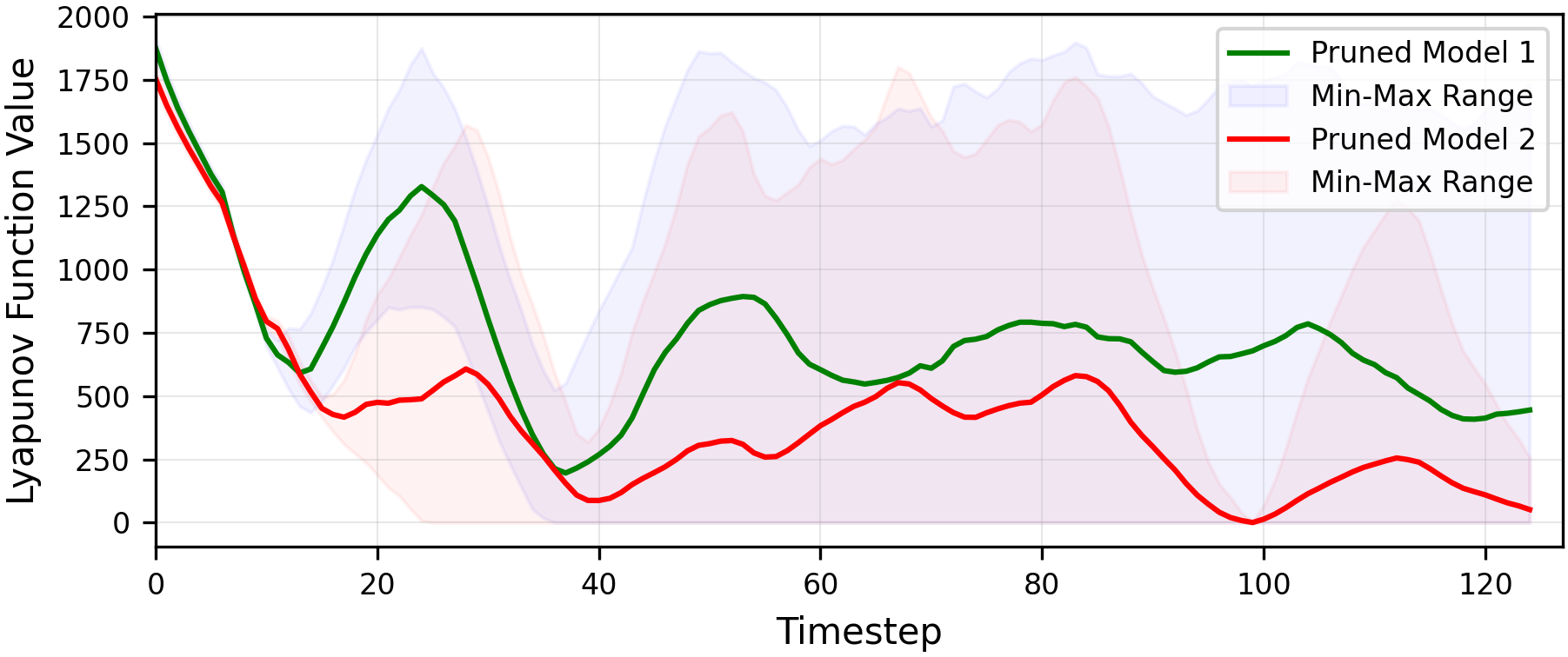}
    \caption{The sensitivity of component-specific pruning on Lyapunov behavior, making it unstable when targeting critical components}
    \label{fig:lyapencoderpruned}
\end{figure}

\vspace{-1em}
While comprehensive pruning experiments established less than 24\% as a stability boundary, a critical discovery emerged that the achievable sparsity depends critically on which specific components are targeted, not just the total pruning percentage. Fig.~\ref{fig:lyapencoderpruned} demonstrates this through targeted component experiments. When groups from the encoder are selectively pruned (Model 1), the system exhibits catastrophic instability despite less than 8\% total sparsity. Similarly, focusing on the group from the dynamics model (Model 2) destabilizes the controller at only 15\% total sparsity, well below the established threshold. Both scenarios exhibit oscillatory Lyapunov function trajectories that fail to converge, thereby clearly violating the stability conditions. This reveals that not all components contribute equally to system stability. The encoder and dynamics components exhibit higher pruning sensitivity, where small capacity reductions propagate instability throughout the control system.

\section{CONCLUSION}
\label{sec:conclusions}

This work addresses a critical challenge in deploying NNCs on embedded hardware, maintaining essential stability guarantees while achieving the aggressive model compression required by resource constraints. Unlike conventional pruning approaches that prioritize accuracy metrics, we introduced a principled framework that treats Lyapunov stability as the primary optimization constraint, fundamentally shifting the pruning paradigm for control applications. Our contribution extends component-aware structured pruning with a novel optimization-based coefficient search methodology that systematically determines optimal pruning ratios for each component group. Through experimental validation on a TD-MPC agent, we established that our framework successfully achieves precise compression targets while maintaining stability guarantees and successfully identifies sharp stability transition boundaries. Most significantly, we revealed that component-specific pruning sensitivity varies dramatically. Critical components like the encoder and dynamics model cause system instability at much lower compression levels, validating the necessity of structured approaches over uniform pruning strategies.

This insight enables safe integration of NNCs into embedded systems while providing practitioners with essential knowledge about stability-critical architectural components. Future work will focus on the earlier identification of sensitive component groups and explore multi-objective optimization, incorporating additional control-theoretic properties, to further enhance this comprehensive framework for deploying neural network controllers in safety-critical embedded applications.

\printbibliography

\end{document}